\documentclass[journal]{IEEEtran}

\usepackage{amsmath}
\usepackage{bm}
\usepackage{cite}
\usepackage{float}
\usepackage{graphicx}
\usepackage{makecell}
\usepackage{siunitx}
\usepackage{url}
\usepackage[hidelinks]{hyperref}

\graphicspath{{figs/}}
\DeclareGraphicsExtensions{.pdf,.png,.jpg,.jpeg}

\newcommand{\tabref}[1]{Table~\ref{#1}}
\newcommand{\figref}[1]{Fig.~\ref{#1}}

\title{\vspace{17.75pt}{\fontsize{16}{19.926}\selectfont\bfseries\spaceskip=5.65pt
Design of a Biomimetic Joint-Covering Skin with Tissue-Like Structure
to Enhance Proprioception in a Musculoskeletal Humanoid\par}\vspace{8.3pt}}

\author{Akihiro Miki$^{1}$, Shun Hasegawa$^{1}$, Yoshimoto Ribayashi$^{1}$, Kento Kawaharazuka$^{1,2}$, and Kei Okada$^{1}$%
\thanks{$^{1}$ The authors are with the Department of Mechano-Informatics, Graduate School of Information Science and Technology, The University of Tokyo, 7-3-1 Hongo, Bunkyo-ku, Tokyo, 113-8656, Japan. {\texttt\small [miki, hasegawa, ribayashi, kawaharazuka, k-okada]@jsk.t.u-tokyo.ac.jp}}%
\thanks{$^{2}$ The author is with the AI Center, Graduate School of Information Science and Technology, The University of Tokyo, Japan.}}

\begin{document}
\maketitle
\vspace*{-32.55pt}
\thispagestyle{empty}
\pagestyle{empty}
\nocite{Miki:2026:JointReceptorsPotential}
\begin{abstract}
    Proprioception in musculoskeletal humanoids is typically estimated primarily from muscle sensing, while the role of cutaneous deformation around joints remains insufficiently explored.
    In biological systems, mechanoreceptors distributed within soft tissue complement muscle feedback and support reliable joint state estimation.
    This study presents the design of a biomimetic joint-covering skin with a tissue-like layered structure that integrates pressure- and stretch-sensitive elements within the joint-covering tissue.
    The proposed skin is implemented on the musculoskeletal humanoid Musashi-W, and its independent proprioceptive capability as well as its integration with muscle sensing are evaluated.
    Experimental results show that the proposed skin alone achieves joint angle estimation with an average error of approximately 3 degrees.
    Furthermore, integration with muscle sensing improves estimation accuracy.
    Owing to its joint-covering structure, the skin may mechanically mitigate the influence of external disturbances on the muscles, and
    the integration of multiple modalities suggests the possibility of contributing to the identification of external stimuli that are difficult to interpret using muscle sensing alone.
    This work presents a design methodology for biomimetic joint-covering skin and demonstrates that such tissue-structured skin can serve as an effective approach for extending proprioceptive systems in musculoskeletal humanoids.

\end{abstract}

\section{Introduction}\label{sec:joint_capsule_introduction}%
  Biomimetic robotics is a research field that aims to improve robotic performance by mimicking the structures and functions of biological systems, which have acquired high adaptability and efficiency through long evolutionary processes.
  Among its targets are musculoskeletal humanoids, which have been developed to achieve natural and flexible motion generation by mimicking human skeletal structures and muscle arrangements \cite{Asano:2016:Kengoro, Kawaharazuka:2019:Musashi}.

  For such robots, accurate perception of body state is indispensable for achieving stable behavior and adaptive motion.
  This perception corresponds to proprioception, the ability to unconsciously sense the position, movement, and loading of one's own body \cite{Proske:2012:ProprioceptiveReview}.
  In conventional robots driven by motors located at joints, joint angles can be directly measured using sensors such as joint encoders.
  However, in musculoskeletal humanoids with complex body structures, direct measurement of joint angles is often difficult.
  Currently, learning-based approaches that estimate joint angles from sensor information such as muscle length and tension have become mainstream \cite{Kawaharazuka:2020:MusculoskeletalAutoEncoder}.
  Nevertheless, proprioceptive

\vfill\newpage

  \begin{figure}[H]
    \vspace{-\intextsep}
    \vspace{-17pt}
    \centering
    \includegraphics[width=0.95\linewidth]{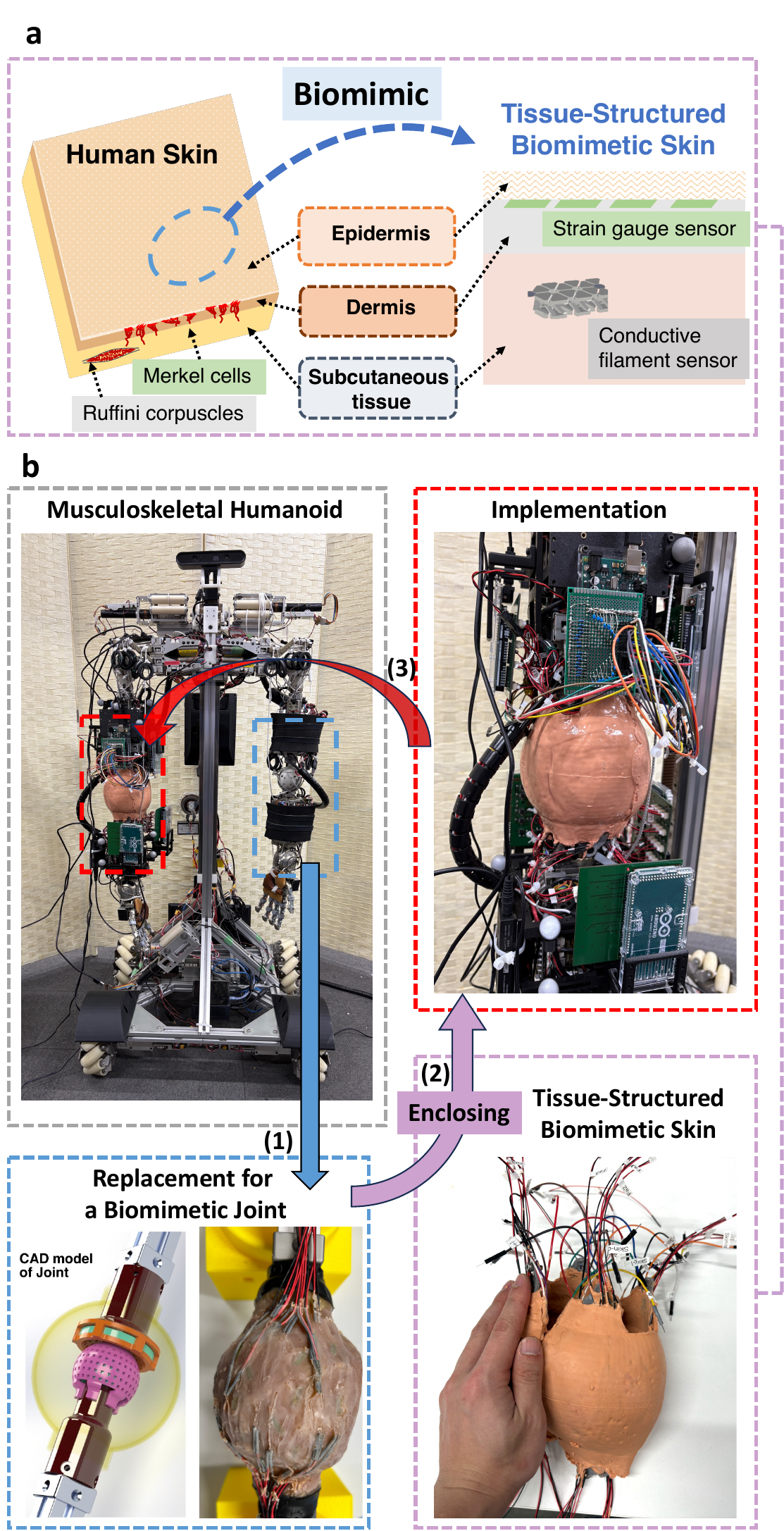}
    \caption{
      Overview of the proposed tissue-structured biomimetic joint-covering skin system.
      \textbf{a.} Concept of the tissue-structured biomimetic skin.
                  The layered structure of biological skin (epidermis, dermis, and subcutaneous tissue) and its mechanoreceptors are mimicked by a multilayer soft structure embedding receptor-like sensors, including strain gauge sensors corresponding to Merkel cells and conductive filament structures corresponding to Ruffini endings.
      \textbf{b.} Implementation on the musculoskeletal humanoid Musashi-W.
                  (1) The conventional joint module is replaced with a biomimetic joint structure adapted from Miki et al. \cite{Miki:2026:JointReceptorsPotential} under the CC BY 4.0 license.
                  (2) The joint region is enclosed by the tissue-structured biomimetic skin.
                  (3) The resulting joint-skin structure is implemented on the musculoskeletal humanoid Musashi-W for enhancing proprioception.
    }
    \label{fig:study_overview}
  \end{figure}

\noindent estimation in many musculoskeletal humanoids still relies on sensors that are limited in both number and modality, such as muscle length sensors and tension sensors.

  In contrast, biological proprioception does not arise from a single receptor but is realized through the integration of afferent signals from numerous mechanoreceptors distributed within soft tissues \cite{Proske:2012:ProprioceptiveReview}.
  In addition to muscle spindles and Golgi tendon organs, receptors located in the skin and around joints are known to contribute to joint state estimation.
  Such sensory organization provides redundancy and contributes not only to improved estimation accuracy but also to discrimination between external disturbances and internal state changes.
  In other words, biological proprioception is established not by the performance of individual receptors alone, but by the framework of tissue organization and information integration.

  In robotics, however, design simplicity and controllability are often prioritized, leading to the use of a small number of high-performance sensors for state estimation \cite{Ni:2019:DynamicParameterIdentification, Saeedvand:2019:HumanoidRobotDevelopment}.
  Studies of proprioception in musculoskeletal humanoids from a biomimetic perspective that incorporates tissue-structured organization of receptor elements and their supporting structures remain limited.

  In this study, we design and implement a biomimetic joint-covering skin that mimics the layered structure of biological skin and embeds sensory elements corresponding to pressure- and stretch-sensitive receptors within soft tissue (\figref{fig:study_overview}(a, b)).
  Unlike conventional approaches that rely on a small number of high-performance sensors, the proposed structure is based on a tissue-structured design concept in which numerous small-scale receptor-like elements are integrated into a tissue-level structural organization.
  The proposed skin is implemented on the musculoskeletal humanoid Musashi-W \cite{Kawaharazuka:2022:MusashiW, Miki:2023:MusashiW}, and its proprioceptive estimation capability as a standalone modality, as well as its integration with a muscle-based sensory system, is evaluated.
  Furthermore, we examine how the integration of muscle-derived and skin-derived information contributes to the interpretation of external stimuli.

\section{Related Works} \label{sec:related_works}
\subsection{Biological Basis of Skin Organization and Proprioception} \label{subsec:related_works_biological_basis}
  Human skin is anatomically composed of three layers: the epidermis, dermis, and subcutaneous tissue \cite{Lotfollahi:2024:SkinAnatomy}.
  Within this layered structure, various types of mechanoreceptors, such as Merkel cells, Pacinian corpuscles, Ruffini endings, and Meissner corpuscles, are distributed at different depths and regions depending on the type of stimulus they detect.
  This spatial organization enables the integration and processing of diverse tactile information within the skin \cite{Johnson:2001:FourTypeMechanoreceptorsRoles, Abraira:2013:SensoryNeuronsTouch}.

  Cutaneous receptors contribute not only to tactile perception but also to proprioception \cite{Proske:2012:ProprioceptiveReview}.
  In particular, slowly adapting type II cutaneous stretch receptors (Ruffini endings) have been suggested to convey information about limb position through skin deformation.
  Thus, multiple types of receptors embedded within the skin participate in both tactile and proprioceptive processing, and their functional roles are closely associated not only with individual receptors but also with the layered tissue structures that support them.

  Based on these anatomical and neurophysiological findings, the present study attempts a biomimetic implementation that incorporates both receptor-like elements and their supporting layered tissue structures.

\subsection{Engineering Approaches to Cutaneous Sensing} \label{subsec:related_works_engineering_approaches}
  Previous studies have proposed tactile sensors inspired by biological systems.
  Examples include BioTac \cite{Fishel:2012:BioTac}, a fluid-filled fingertip tactile sensor with multiple electrodes, and an anthropomorphic fingertip sensor embedding strain gauges and PVDF films in an elastic body \cite{Hosoda:2006:AnthropomorphicFinger}.
  These sensors have been primarily developed for object recognition and contact-state estimation rather than for constructing proprioception.

  Modular skin systems designed to cover the entire robot body have also been proposed \cite{Ohmura:2006:ConformableSkin, Cheng:2019:ComprehensiveSkin}.
  Although these systems are important for realizing wide-area tactile distribution, their primary objective is contact detection and tactile mapping, and proprioception is not their main focus.

  Electronic skin (e-skin) technologies that integrate numerous sensors on flexible substrates \cite{Zhang:2024:ElectronicSkinReview} represent another promising approach for acquiring large-scale sensory information.
  However, their main focus remains tactile discrimination and high-density sensing.
  When extending planar substrates to three-dimensional joint geometries, practical challenges remain, including implementation complexity and potential effects on joint mechanical properties.

  There also exist engineering approaches that estimate body states using sensors embedded within or attached to the body.
  In soft robotics, embedded sensors have been used for shape reconstruction \cite{Truby:2020:DistributedSoftSensor}, and sensor integration has been explored in soft arms inspired by octopus morphology \cite{Soter:2018:OctopusProprioceptionArm}.
  In these studies, sensors are deliberately arranged to achieve specific functional goals, and the underlying design philosophy typically aims to accomplish tasks with a minimal number of informative sensors.

  In contrast, biological sensory receptors are not arranged as engineering-style grids or optimally designed layouts, but instead exist in large numbers within organized tissue structures.
  Motivated by this perspective, the present study does not optimize sensor placement from the initial stage of body construction.
  Instead, it adopts as a design principle the incorporation of numerous receptor-like elements together with their supporting tissue-like structures.

  Such a body construction strategy is expected to enable a proprioceptive architecture that does not rely on a small number of high-performance sensors.
  Moreover, by constructing a body that resembles human structural organization, this approach is consistent with the constructive perspective emphasized in biomimetics, which seeks to derive insights through reproducing biological structural principles.

  Overall, compared with engineering studies that primarily pursue optimized bodies for functional realization, this work is positioned as an attempt to mimic biological complexity at the tissue level and to examine the functions that emerge from such structural organization.

\section{Development of the Tissue-Structured Biomimetic Joint-Covering Skin} \label{sec:design_and_implementation}
  Unlike conventional engineering approaches that optimally place a small number of high-performance sensors,
  this study adopts a biomimetic design principle based on biological skin, where numerous receptors are embedded within a layered tissue structure that mechanically supports them.

  \begin{figure}[htbp]
    \centering
    \includegraphics[width=\linewidth]{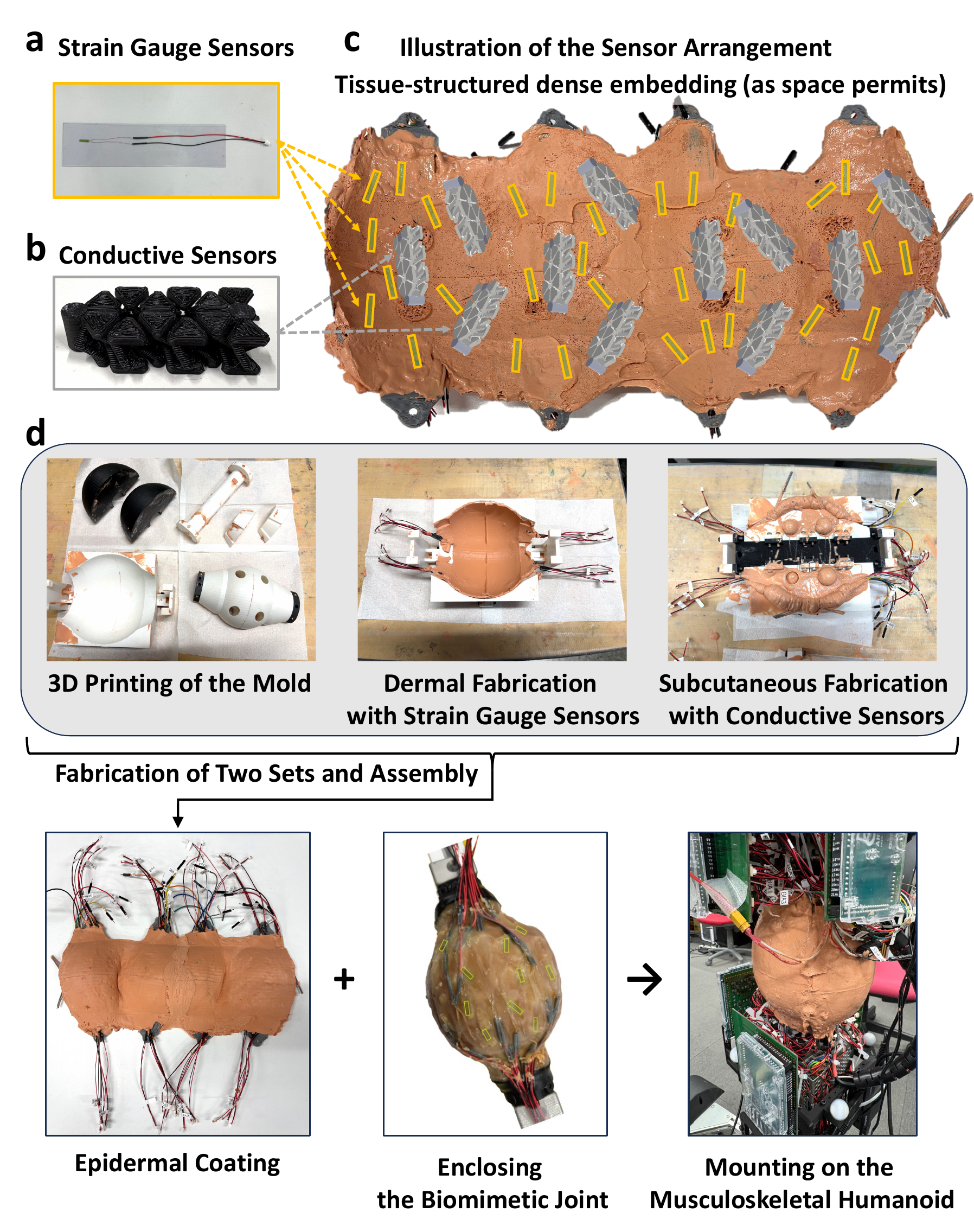}
    \caption{
      Fabrication and integration process of the tissue-structured biomimetic joint-covering skin.
      \textbf{a.} Strain gauge sensors used as Merkel cell-like elements.
      \textbf{b.} Conductive chainmail sensors used as stretch-sensitive elements corresponding to Ruffini endings.
      \textbf{c.} Illustration of the sensor arrangement in the biomimetic skin, showing the spatial distribution of strain gauge sensors and conductive chainmail sensors.
                  Sensors are placed with reference to biological tissue organization and embedded as densely as space permits.
      \textbf{d.} Fabrication procedure: 3D printing of the mold, embedding strain gauge sensors during dermal layer fabrication, and embedding conductive chainmail sensors during subcutaneous layer fabrication.
                  A pair of skin components is then fabricated and combined.
                  The epidermal layer is applied by thin coating, the biomimetic joint is enclosed, and the structure is finally mounted onto the musculoskeletal humanoid.
                  The biomimetic joint is adapted from Miki et al. \cite{Miki:2026:JointReceptorsPotential} under the CC BY 4.0 license.
    }
    \label{fig:biomimetic_skin_fabrication_details}
  \end{figure}

\subsection{Tissue-like Layered Structure and Materials} \label{subsec:design_and_implementation_tissue_structure}

  The biomimetic skin used in this study was constructed as a three-layer structure corresponding to the epidermis, dermis, and subcutaneous tissue.
  Human skin is a composite tissue that exhibits viscoelasticity, anisotropy, and nonlinear mechanical behavior, making it difficult to strictly reproduce its mechanical properties using a single indicator.
  Therefore, rather than aiming at a complete reproduction of biological skin, this study focused on mimicking the relative mechanical tendency that compliance increases from the outer layer to the inner layers.
  The material composition was selected with reference to our previous work \cite{Miki:2026:BiomimeticSkinComparison}, which investigated the configuration and physical properties of a biomimetic soft-tissue structure.

  The materials used for each layer are as follows:
  \begin{itemize}
  \item Epidermis: Dragon Skin 30 (silicone rubber, Shore hardness 30A, Smooth-On, Inc.)
  \item Dermis: Ecoflex Gel 2 (silicone gel, Shore hardness 000-34, Smooth-On, Inc.)
  \item Subcutaneous tissue: Soma Foama 15 (foamed silicone sponge, \SI{240}{\kilogram\per\cubic\meter}, Smooth-On, Inc.)
  \end{itemize}

  The thickness of biological skin layers varies with individuals and body location.
  Previous studies report that the epidermis is typically tens to hundreds of micrometers thick, the dermis several millimeters thick, and the subcutaneous tissue ranges from several millimeters to several centimeters \cite{McGrath:2016:GrayAnatomy, Sim:2014:AppropriatenessInsulinNeedles}.
  Referring to these anatomical values, the fabricated skin was designed with approximate layer thicknesses of \SI{0.2}{\milli\meter} for the epidermis, \SI{2}{\milli\meter} for the dermis, and \SI{10}{\milli\meter} for the subcutaneous layer.
  Each layer was colored using Silc Pig Light Flesh (silicone pigment, Smooth-On, Inc.) to facilitate visual identification during fabrication.

\subsection{Receptor-like Elements} \label{subsec:design_and_implementation_receptor_elements}
  Ideally, all types of cutaneous receptors in biological skin would be reproduced.
  However, due to spatial constraints within the embedded structure, this study focuses on selected mechanoreceptors.
  For proprioception, slowly adapting type II (SA2) afferents, often associated with Ruffini endings and skin stretch detection, have been suggested to contribute to joint position sensing.
  Therefore, a sensor structure mimicking Ruffini endings was incorporated.
  In addition, Merkel cells, which are connected to slowly adapting type I (SA1) afferents and respond to sustained pressure, were also selected due to their similar adaptation characteristics.

  For the Merkel cell-like elements, strain gauge sensors (General-purpose Foil Strain Gages, KYOWA ELECTRONIC INSTRUMENTS CO., LTD., Tokyo, Japan) were employed (\figref{fig:biomimetic_skin_fabrication_details}(a)), whose electrical resistance changes in response to applied strain.
  Each strain gauge has dimensions of \SI{5}{\milli\meter} in length, \SI{1.4}{\milli\meter} in width, and \SI{14}{\micro\meter} in thickness, and was connected to an Arduino MEGA via a voltage divider circuit for signal acquisition.

  Ruffini endings are morphologically described as spindle-shaped encapsulated structures that respond to skin stretch, although their detailed transduction mechanisms remain incompletely understood.
  In this study, a stretch-responsive sensor structure was constructed using a conductive filament (Conductive Filaflex, Recreus Industries S.L., Alicante, Spain) arranged in a chainmail configuration.
  The structure was designed such that its electrical resistance changes in response to in-plane elongation of the surrounding tissue.
  The chainmail structure was fabricated by assembling multiple claw-connected triangular units with a distance of \SI{2.5}{\milli\meter} from the center to each edge, followed by the attachment of electrode terminals.
  The resulting structure (thickness \SI{8}{\milli\meter}, length \SI{36}{\milli\meter}, width \SI{22}{\milli\meter}) was embedded within the biomimetic skin, as shown in \figref{fig:biomimetic_skin_fabrication_details}(b).

  Referring to the depth-dependent distribution of biological receptors, the Merkel cell-like strain gauges were embedded between the epidermal and dermal layers, while the Ruffini-like conductive structures were embedded within the subcutaneous layer.
  By embedding these receptor-like elements into the tissue-structured biomimetic skin, depth-specific sensory information corresponding to each receptor type can be obtained and evaluated.

\subsection{Development of Tissue-Structured Biomimetic Skin and Its Implementation in a Musculoskeletal Humanoid} \label{subsec:development_and_implementation}
  Anatomically, the dermis and subcutaneous tissue are connected to the fascia via fibrous structures known as skin ligaments \cite{Nash:2004:SkinLigamentsRegionalDistribution}.
  Through this structural continuity, deformation of the skin can be transmitted to the skeletal system.
  Based on these anatomical observations, bundled thermoplastic polyurethane (TPU95A-HF, Bambu Lab Inc., Shenzhen, China) fibers were embedded within the subcutaneous layer to mimic continuous collagen fiber structures.
  During fabrication, the embedded fiber bundles were arranged to achieve mechanical continuity with the skeletal structure of the musculoskeletal humanoid.

  Details of the fabrication process are shown in \figref{fig:biomimetic_skin_fabrication_details}(c, d).
  Although multilayer structures are typically fabricated in the order of epidermis-dermis-subcutaneous tissue or the reverse, forming the epidermis at a thickness of approximately \SI{0.2}{\milli\meter} resulted in fracture during mold removal after curing.
  Conversely, fabricating the subcutaneous layer first made it difficult to uniformly form the upper layers due to the presence of micro-pores in the foamed silicone sponge.
  Therefore, the dermis and subcutaneous layers were first fabricated, followed by thin coating of the epidermal layer.
  The dermis and subcutaneous layers were formed with approximate thicknesses of \SI{2}{\milli\meter} and \SI{10}{\milli\meter}, respectively, while the epidermis was formed by thin coating.

  During fabrication of the dermal layer, strain gauge sensors corresponding to Merkel cell-like elements were attached to the mold wall, and Ecoflex Gel 2 was poured and cured to position the sensors at the epidermis-dermis interface.
  A total of 32 sensors were embedded within the tissue.
  Rather than arranging them in a regular grid, the sensors were distributed in a semi-random manner, referring to biological receptor distributions.

  Subsequently, the subcutaneous layer was formed by replacing the core and pouring Soma Foama 15 onto the dermal layer.
  During this process, 12 conductive chainmail structures corresponding to Ruffini-like elements, along with the fiber structures, were embedded.
  Since Ruffini endings are reported to align along collagen fibers within the skin, the chainmail structures were oriented along the direction of the embedded fiber bundles.

  After demolding, Dragon Skin 30 was thinly coated and cured to form the epidermal layer.
  This process was repeated twice to produce a pair of biomimetic skin components.

  The fabricated skin pair was used to envelop an open-structure biomimetic joint, in which bones are connected by compliant tissue \cite{Miki:2026:JointReceptorsPotential}, and mounted onto the musculoskeletal humanoid.
  The two skin components were connected by applying Soma Foama 15 at their interface.

  The musculoskeletal humanoid Musashi-W used in this study is tendon-driven, and each muscle module measures muscle length and tension.
  These muscle-derived signals are used as proprioceptive information corresponding to muscle spindles and Golgi tendon organs in biological systems.
  In this configuration, joint-covering skin information is integrated with this muscle-derived proprioceptive information.

\section{Experiments using Tissue-Structured Skin and Muscle Sensing} \label{sec:experiments}
\subsection{Sensorimotor Data Collection} \label{subsubsec:sensorimotor_data_collection}
  Sensorimotor data were collected while operating the musculoskeletal humanoid equipped with the tissue-structured biomimetic skin, and the acquired data were used to evaluate proprioceptive estimation.

  Data acquisition focused on the right upper limb.
  In Musashi-W, five pairs of muscle modules are arranged to actuate the shoulder and elbow joints.
  Each muscle module provides measurements of muscle length (corresponding to muscle spindle sensing) and muscle tension (corresponding to Golgi tendon organ sensing).
  Accordingly, muscle-derived proprioceptive information was obtained from 10 muscle length sensors and 10 tension sensors.

  In addition to these muscle signals, sensory information from the tissue-structured biomimetic skin developed in this study was also recorded.
  The skin sensory system consists of 32 strain gauge sensors corresponding to Merkel-cell-like receptors and 12 stretch sensors corresponding to Ruffini-like receptors.
  These muscle-derived and skin-derived signals were used to estimate the elbow joint angles.

  In the original Musashi-W system, joint angle sensors (potentiometers) were embedded inside the joint modules.
  However, in this study, the joint modules were replaced with tissue-biomimetic joints, and the potentiometers were not used.
  Instead, motion capture markers were attached to the upper arm and forearm, and the three-dimensional pose was measured using the motion capture system Motive (version 3.0.1, OptiTrack, NaturalPoint, Inc.).
  The elbow joint angles (pitch, yaw, and roll) were computed from the relative transformation between the upper arm and forearm using the TF package in ROS based on their quaternion orientations.
  These joint angles were used as the ground-truth signals for proprioceptive estimation.

\begin{figure}[htbp]
    \centering
    \includegraphics[width=\linewidth]{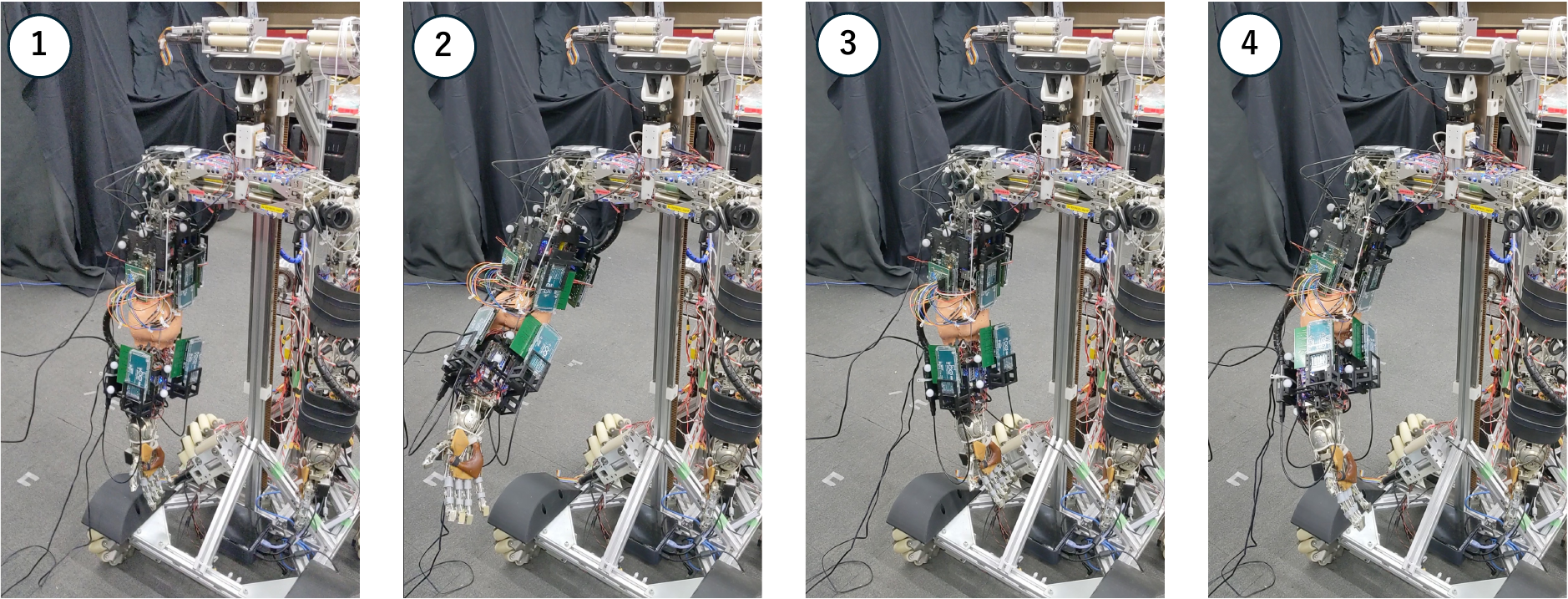}
    \caption{
      Process of sensorimotor data collection using the musculoskeletal humanoid equipped with the tissue-structured biomimetic joint-covering skin.
      During robot motion, muscle-related signals (muscle length and muscle tension), cutaneous signals from distributed strain gauges (Merkel-cell-like sensors) and conductive sensors (Ruffini-like sensors), and motion capture measurements are recorded simultaneously.
      These signals are used to construct a sensorimotor dataset for evaluating proprioceptive estimation.
    }
    \label{fig:data_collection_process}
  \end{figure}

  \figref{fig:data_collection_process} illustrates the data acquisition process.
  Sensorimotor data were collected while moving the right upper limb through a variety of postures.
  Motion generation was performed using the body schema model proposed in previous studies \cite{Kawaharazuka:2019:Musashi, Kawaharazuka:2022:MusashiW}.
  Because the body structure was modified from the original robot, the original model did not guarantee precise joint-angle control.
  Nevertheless, it provided sufficient posture variation for constructing the sensorimotor dataset used in this study.
  Furthermore, the tissue-biomimetic joint allows a roll degree of freedom at the elbow, enabling a wider range of postures than the original joint module.

  Data collection was performed by repeatedly generating a motion command, allowing the robot to move to the commanded posture, waiting until the posture stabilized, and recording the sensor values.
  This procedure was repeated for approximately one hour of operation, resulting in a dataset consisting of 615 sensorimotor samples.
  The collected dataset covered approximately \SI{50}{\degree} of elbow motion along each of the pitch, yaw, and roll axes.
  This range does not represent the mechanical limit of the joint, but was selected as an experimental operating range to avoid disconnection of the embedded sensors or changes in the number of valid sensor channels, thereby enabling reproducible evaluation.

\subsection{Single-Modality Proprioceptive Estimation} \label{subsec:singlemodal_proprioceptive_estimation}
  To compare proprioceptive estimation performance across individual modalities,
  three-degree-of-freedom elbow joint angles were estimated using Muscle Length (ML), Muscle Tension (MT), Ruffini-like Skin Sensor (SKIN-R), and Merkel-like Skin Sensor (SKIN-M).

  The collected sensorimotor dataset was shuffled and split into training (70\%), validation (15\%), and test (15\%) sets using 20 random seeds.
  Input features were standardized using the mean and standard deviation computed from the training set and applied to the validation and test sets.

  A fully connected multilayer perceptron (MLP) was used as the estimation model.
  It consisted of three hidden layers with 256 units each, ReLU activation functions, dropout (0.1), and a linear output layer predicting three joint angles.

  Model training used the AdamW \cite{Loshchilov:2017:AdamW} optimizer with a learning rate of $3\times10^{-4}$, weight decay of $1\times10^{-4}$, and mean squared error (MSE) loss.
  The batch size was 256, and the maximum number of epochs was 400.

\begin{figure}[htbp]
    \centering
    \includegraphics[width=\linewidth]{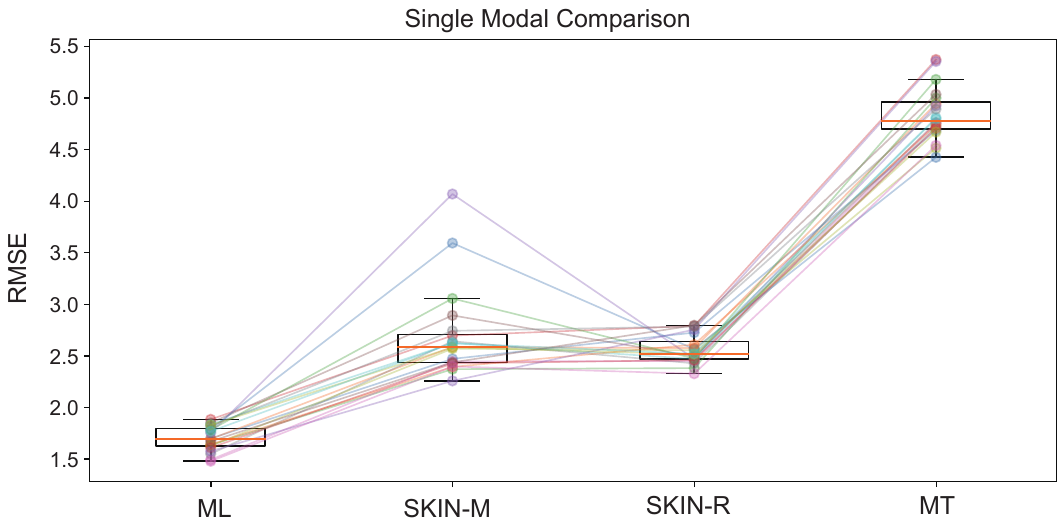}
    \caption{
      Comparison of single-modality joint angle estimation performance for the elbow (3DOF).
      RMSE are averaged across the three joint angles and paired by random seed ($n=20$).
      ML: muscle length, SKIN-R: Ruffini-like skin sensor, SKIN-M: Merkel-like skin sensor, MT: muscle tension.
    }
    \label{fig:rmse_single_modal}
  \end{figure}

  Estimation performance was evaluated using root mean squared error (RMSE), computed for each joint-angle dimension and averaged across dimensions.
  The RMSE distribution for each modality is shown in \figref{fig:rmse_single_modal}, and summary statistics across the 20 random seeds are shown in \tabref{tab:single_modal_summary}.

  For statistical comparisons between modalities, RMSE values obtained from the 20 random seeds were compared as paired data.
  Since the normality could not be reliably assumed, the Wilcoxon signed-rank test was applied with Holm correction for multiple comparisons.
  The results are summarized in \tabref{tab:single_modal_stats}.

  The muscle length modality exhibited the smallest estimation error in terms of both mean and median RMSE, and statistically significant differences were observed between this modality and the others (Holm-corrected $p < 0.001$).
  Although the skin modalities showed larger errors than the muscle length modality, their RMSE values were generally distributed within a range of approximately three degrees.

  \begin{table}[htbp]
    \centering
    \caption{Summary of elbow joint angle estimation errors for single-modality conditions.}
    \label{tab:single_modal_summary}
    \begin{tabular}{lcccc}
      \hline
      Modality & $n$ & Mean RMSE & SD & Median RMSE \\
      \hline
      Muscle length (ML) & 20 & 1.70 & 0.12 & 1.70 \\
      \makecell{Ruffini-like Skin Sensor \\ (SKIN-R)} & 20 & 2.56 & 0.14 & 2.52 \\
      \makecell{Merkel-like Skin Sensor \\ (SKIN-M)} & 20 & 2.69 & 0.44 & 2.58 \\
      Muscle tension (MT) & 20 & 4.84 & 0.25 & 4.78 \\
      \hline
    \end{tabular}
  \end{table}

  \begin{table}[htbp]
    \centering
    \caption{Paired statistical comparison between muscle length and other modalities (Wilcoxon signed-rank test, $n=20$).}
    \label{tab:single_modal_stats}
    \begin{tabular}{lccc}
      \hline
      Comparison & Mean difference (ML -- X) & $p$ & $p_{\mathrm{Holm}}$ \\
      \hline
      ML vs MT & $-3.14$ & $<10^{-5}$ & $<0.001$ \\
      ML vs SKIN-M & $-0.99$ & $<10^{-5}$ & $<0.001$ \\
      ML vs SKIN-R & $-0.86$ & $<10^{-5}$ & $<0.001$ \\
      \hline
    \end{tabular}
  \end{table}

\subsection{Multimodal Proprioceptive Estimation} \label{subsec:multimodal_proprioceptive_estimation}
  To evaluate the effect of integrating muscle-derived and skin-derived information on proprioceptive estimation,
  joint angle estimation experiments using multimodal sensory inputs were conducted.
  Three modality combinations were defined:
  muscle-related information combining ML and MT (M\_fused), skin-related information combining SKIN-M and SKIN-R (T\_fused), and a multimodal condition combining both muscle and skin information (M+T).

  Two architectures were compared for modality fusion.
  The first was a concatenation-based multilayer perceptron (concat), in which features from different modalities were directly concatenated and fed into a single MLP.
  The network consisted of three hidden layers with 256 units each, using ReLU activation functions.

  The second was an encoder-based fusion model (enc), in which each modality was processed by an encoder and projected into a low-dimensional embedding space.
  Each encoder consisted of two layers with 128 units, producing a 32-dimensional embedding.
  The embeddings were concatenated and passed to a regression head composed of two hidden layers with 256 units each to predict the joint angles.
  ReLU activation functions were used throughout the network.

\begin{figure}[htbp]
    \centering
    \includegraphics[width=\linewidth]{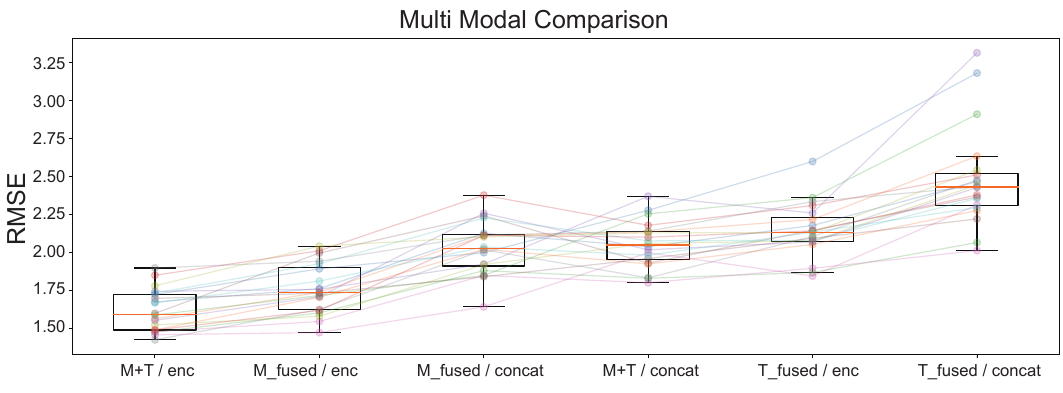}
    \caption{
      Comparison of multimodal proprioception estimation performance for the elbow (three degrees of freedom).
      RMSE values are averaged across targets and paired by random seed ($n=20$).
      Results are shown for different modality combinations and fusion architectures.
    }
    \label{fig:rmse_fusion}
  \end{figure}

  The procedures for dataset splitting, input normalization, training conditions, and statistical analysis were identical to those used in the single-modality experiments.
  Estimation performance under multimodal conditions was evaluated using RMSE values from 20 random seeds, following the same procedure as in the single-modality experiments.
  The distribution of RMSE for each modality combination and fusion architecture is shown in \figref{fig:rmse_fusion}.
  Statistical comparisons were performed using ML as the baseline, and the Wilcoxon signed-rank test was applied to paired RMSE values with Holm correction for multiple comparisons.
  Summary statistics are shown in \tabref{tab:fusion_modal_summary}, and the statistical comparison results are presented in \tabref{tab:multimodal_stats}.

  The encoder-based fusion of muscle and skin modalities (M+T\_enc) achieved higher estimation accuracy than the muscle-length-only condition, and the difference was statistically significant (Holm-corrected $p < 0.001$).
  In contrast, the other fusion conditions did not consistently outperform the muscle-length-only condition.

  \begin{table}[htbp]
    \centering
    \caption{Summary of elbow joint angle estimation errors for multimodal fusion conditions.}
    \label{tab:fusion_modal_summary}
    \begin{tabular}{lcccc}
      \hline
      Method & $n$ & Mean RMSE & SD & Median RMSE \\
      \hline
      M\_fused (concat) & 20 & 2.03 & 0.19 & 2.01 \\
      M\_fused (enc) & 20 & 1.76 & 0.14 & 1.74 \\
      T\_fused (concat) & 20 & 2.48 & 0.22 & 2.45 \\
      T\_fused (enc) & 20 & 2.15 & 0.20 & 2.12 \\
      M+T (concat) & 20 & 2.05 & 0.17 & 2.02 \\
      \textbf{M+T (enc)} & 20 & \textbf{1.62} & 0.11 & 1.60 \\
      \hline
    \end{tabular}
  \end{table}

  \begin{table}[htbp]
    \centering
    \caption{Paired statistical comparison between the muscle length modality and multimodal fusion conditions (Wilcoxon signed-rank test, $n=20$).}
    \label{tab:multimodal_stats}
    \begin{tabular}{lccc}
      \hline
      Comparison
      & \makecell{Mean diff. \\ (ML--Fusion)}
      & $p$
      & $p_{\mathrm{Holm}}$ \\
      \hline
      ML vs M\_fused (concat) & $-0.334$ & $2.0\times10^{-6}$ & $1.1\times10^{-5}$ \\
      ML vs T\_fused (concat) & $-0.781$ & $2.0\times10^{-6}$ & $1.1\times10^{-5}$ \\
      ML vs T\_fused (enc)    & $-0.446$ & $2.0\times10^{-6}$ & $1.1\times10^{-5}$ \\
      ML vs M+T (concat)      & $-0.353$ & $2.0\times10^{-6}$ & $1.1\times10^{-5}$ \\
      ML vs M\_fused (enc)    & $-0.061$ & $1.0\times10^{-3}$ & $1.0\times10^{-3}$ \\
      ML vs M+T (enc)         & $\mathbf{+0.077}$ & $3.2\times10^{-4}$ & $6.5\times10^{-4}$ \\
      \hline
    \end{tabular}
  \end{table}

\subsection{Evaluation of Joint-Covering Skin under External Mechanical Disturbances} \label{subsec:external_disturbances_experiment}
  In many existing musculoskeletal humanoids, the muscles responsible for driving the body are exposed to the external environment.
  As a result, external mechanical stimuli may be directly applied to the muscles.
  When proprioception is estimated using only muscle-derived signals, it is difficult to determine whether changes in sensor readings originate from external stimulation or from changes in muscle state.
  The proposed joint-covering skin surrounds the joint region and can mechanically mediate external stimuli while also providing additional sensory information.
  By integrating skin- and muscle-derived information, external mechanical disturbances may be interpreted more reliably than when using muscle-derived sensing alone.

\begin{figure}[htbp]
    \centering
    \includegraphics[width=\linewidth]{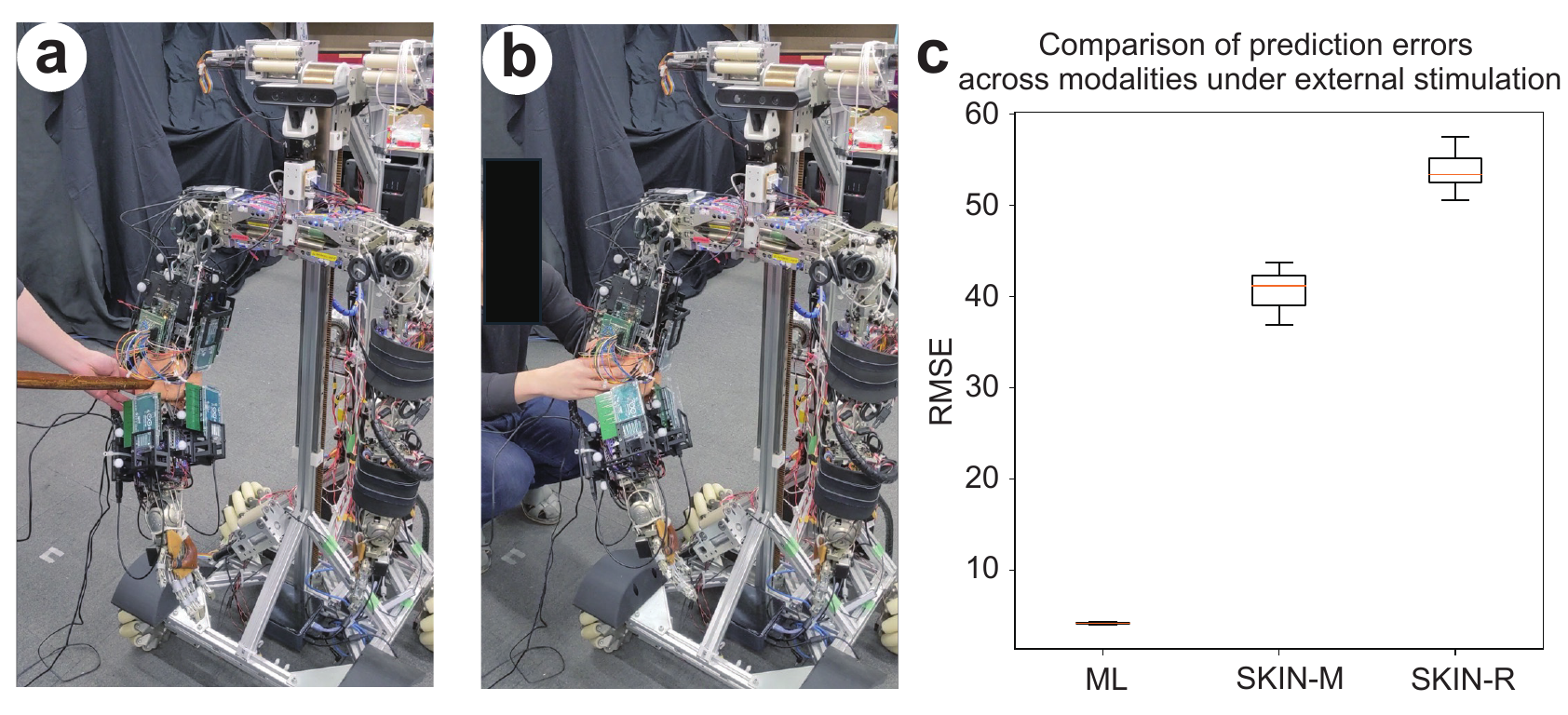}
    \caption{
      External mechanical stimulation experiment.
      \textbf{a.} External perturbation applied to the elbow region using a rod during robot motion.
      \textbf{b.} External perturbation applied manually by hand.
      \textbf{c.} Comparison of joint angle estimation errors (RMSE) across sensing modalities under external stimulation.
      RMSE values are averaged across the three elbow joint angles and paired by random seed ($n=20$).
      ML: muscle length, SKIN-M: Merkel-like skin sensor, SKIN-R: Ruffini-like skin sensor.
    }
    \label{fig:sensorimotor_data_collection_with_stimulus_and_boxplot}
  \end{figure}

  To investigate this role of the joint-covering skin, additional sensorimotor data were collected while external stimuli were applied around the joint during robot motion.
  The data collection process is illustrated in \figref{fig:sensorimotor_data_collection_with_stimulus_and_boxplot} (a, b).
  External disturbances were introduced by pressing the elbow joint region with a stick or by applying pressure by hand from various directions.
  Sensorimotor data were recorded under these disturbance conditions for approximately 15 minutes, resulting in 157 sensorimotor samples.

  Using these data, the proprioceptive estimation performance of the muscle and skin modalities was evaluated separately under single-modality conditions.
  The training conditions and estimation procedures were identical to those used in the single-modality experiments described in the previous section.

  Models trained under the no-disturbance condition were used to estimate joint angles from sensor data obtained under external disturbance conditions.
  The results are shown in \figref{fig:sensorimotor_data_collection_with_stimulus_and_boxplot} (c) and \tabref{tab:external_disturbance_stats}.
  Under disturbance conditions, estimation errors increased for both the skin and muscle modalities, with particularly large degradation observed for the skin modality.
  The estimation error of the muscle modality also increased under disturbance conditions; however, the mean error remained approximately four degrees, indicating that the influence of external mechanical stimuli on muscle signals was relatively limited.

  \begin{table}[htbp]
    \centering
    \caption{Effect of external mechanical stimulation on elbow joint angle estimation error (RMSE, deg).}
    \label{tab:external_disturbance_stats}
    \begin{tabular}{lccc}
      \hline
      Modality & No-stim & Stim & $p_{\mathrm{Holm}}$ \\
      \hline
      Muscle length (ML) & $1.70\pm0.12$ & $4.18\pm0.07$ & $<0.001$ \\
      \makecell{Ruffini-like Skin Sensor \\ (SKIN-R)} & $2.56\pm0.14$ & $53.63\pm1.82$ & $<0.001$ \\
      \makecell{Merkel-like Skin Sensor \\ (SKIN-M)} & $2.69\pm0.44$ & $40.81\pm1.99$ & $<0.001$ \\
      \hline
    \end{tabular}
  \end{table}

\section{Discussion} \label{sec:discussion}
  The results of Single-Modality Proprioceptive Estimation (\ref{subsec:singlemodal_proprioceptive_estimation}) showed that although the skin modality exhibited larger errors than the muscle length modality,
  joint angle estimation was still possible with an accuracy of approximately three degrees.
  This result suggests that deformation of the skin surrounding the joint contains information related to proprioception.
  It is also consistent with biological findings indicating that cutaneous mechanoreceptors, such as Ruffini endings, contribute to proprioception through skin stretch.

  The results of Multimodal Proprioceptive Estimation (\ref{subsec:multimodal_proprioceptive_estimation}) showed that the encoder-based fusion of muscle and skin modalities (M+T\_enc) achieved higher estimation accuracy than the muscle-length-only condition.
  This finding suggests that muscle-derived and skin-derived signals contain complementary information and that integrating these modalities through appropriate representation learning can enable more accurate joint angle estimation than single-modality sensing.

  The results of the external disturbance experiment (\ref{subsec:external_disturbances_experiment}) showed that estimation errors increased in both muscle and skin modalities when external mechanical stimuli were applied.
  In particular, the skin modality was strongly affected because local deformation caused by external contact was superimposed on the deformation produced by joint motion, leading to a substantial shift in the input distribution relative to the models trained under non-stimulated conditions.
  At the same time, the modality-dependent change in estimation performance may provide a cue for interpreting external mechanical disturbances around the joint.
  For example, if skin-derived estimates deviate substantially from muscle-derived estimates or from the expected joint state, while muscle-derived estimates remain relatively stable, this may indicate external contact around the joint region.
  Conversely, if skin-derived estimates remain relatively stable while muscle-derived estimates change substantially, this may indicate a muscle-side abnormality, such as abnormal changes in muscle tension, tendon slack, or failure of the muscle sensing system, rather than external contact around the joint.
  Such interpretation is enabled by comparing the estimation results from the muscle and skin modalities; when only the muscle modality is used, it is difficult to determine whether changes in sensor outputs or estimation results are caused by external stimuli or by changes in the musculoskeletal state.
  Furthermore, the skin structure covering the joint may also function as a mechanical protective layer that mitigates the direct transmission of external stimuli to the muscles.
  Taken together, these results suggest that the proposed tissue-structured biomimetic skin may contribute to improving the reliability of proprioceptive systems by providing complementary sensory information during multimodal integration,
  as well as by supporting disturbance interpretation and providing mechanical protection.

  The joint-covering skin structure proposed in this study may be useful not only from a biomimetic perspective but also for general robotic systems.
  By integrating distributed sensors within a compliant skin structure, it may be possible to simultaneously provide mechanical protection, contact detection, and complementary sensory information.
  Such characteristics could be advantageous not only for musculoskeletal humanoids but also for a wide range of robots that physically interact with their environments.

  However, several limitations remain in the present study.
  The receptor-like sensors used in this work are significantly larger than biological mechanoreceptors, and the number of sensors is substantially smaller than that found in biological skin.
  Therefore, the proposed implementation should be interpreted not as a strict reproduction of biological structures but rather as a conceptual reconstruction of a tissue-level sensory architecture.
  Future work should improve sensor miniaturization and density to realize sensory structures that more closely resemble biological systems.

  In addition, the long-term stability of the proposed skin and its response during high-speed motion were not fully evaluated in this study.
  Due to the viscoelasticity and history-dependent behavior of silicone-based materials, hysteresis, fatigue, and drift may occur during long-term use and affect estimation accuracy.
  Furthermore, because the dataset used in this study was mainly collected under quasi-static conditions, evaluating performance during fast and dynamic motions, as well as introducing online calibration, adaptive learning, and time-series models, remains future work.

\section{Conclusion} \label{sec:conclusion}
  This study presented the design and implementation of a biomimetic joint-covering skin with a tissue-like structure aimed at enhancing proprioception in a musculoskeletal humanoid.
  The proposed structure is based on a tissue-level design approach that mimics biological sensory organization, in which numerous small-scale receptor-like elements are integrated into soft tissue rather than relying on a small number of high-performance sensors commonly adopted in robotics.
  Experimental results demonstrated that the proposed approach enables meaningful joint angle estimation using only receptor-like elements embedded within the tissue structure, without employing high-precision single sensors or extensive optimization to maximize their performance.
  The results also showed that skin-derived information contributes to improving estimation accuracy when integrated with muscle sensing.
  Furthermore, the flexible skin structure covering the joint may function as a mechanical protective structure that mitigates the direct transmission of external stimuli to the muscles, and may also contribute to the interpretation of external mechanical stimuli.
  The structure of the biomimetic skin and the concept of multisensory integration demonstrated in this study may be useful not only for biomimetic robotics but also for general robotic systems.
  In addition,
  the significance of this study lies in its exploration of biomimicry at the tissue level, focusing on the organized incorporation of numerous receptor-like elements and their supporting structures rather than merely mimicking isolated functions.
  Although this study focused on a tissue-structured joint-covering skin, the same tissue-level design approach may be extended to other body regions, potentially contributing to the development of humanoid robots with more human-like structural characteristics.

{
  \bibliographystyle{IEEEtran}
  \bibliography{references}
}

\end{document}